\documentclass{IOS-Book-Article}

\usepackage{mathptmx}
\usepackage{graphicx}
\usepackage{layouts}

\begin{document}
\begin{frontmatter}
\title{Negotiating Control: Neurosymbolic~Variable Autonomy}
\runningtitle{Neurosymbolic Variable Autonomy}

\author[A]{\fnms{Georgios} \snm{Bakirtzis}},
\author[B]{\fnms{Manolis} \snm{Chiou}}
\and
\author[A]{\fnms{Andreas} \snm{Theodorou}}

\runningauthor{G. Bakirtzis et al.}
\address[A]{Universitat Politècnica de Catalunya}
\address[B]{Queen Mary University of London}

\begin{abstract}
Variable autonomy equips a system, such as a robot, with mixed initiatives such that it can adjust its independence level based on the task's complexity and the surrounding environment. Variable autonomy solves two main problems in robotic planning: the first is the problem of humans being unable to keep focus in monitoring and intervening during robotic tasks without appropriate human factor indicators, and the second is achieving mission success in unforeseen and uncertain environments in the face of static reward structures. An open problem in variable autonomy is developing robust methods to dynamically balance autonomy and human intervention in real-time, ensuring optimal performance and safety in unpredictable and evolving environments. We posit that addressing unpredictable and evolving environments through an addition of rule-based symbolic logic has the potential to make autonomy adjustments more contextually reliable and adding feedback to reinforcement learning through data from mixed-initiative control further increases efficacy and safety of autonomous behaviour.\end{abstract}

\end{frontmatter}

\thispagestyle{empty}
\pagestyle{empty}

\section{Introduction}
Environmental noise makes real-time planning a demanding computational task, adding uncertainty to decision-making.
As a solution, in safety-critical missions conducted in extreme environments (for example, underwater), full teleoperation is still being used; foregoing the benefits of automation.
Even where automation is used, human oversight is still required.
However, multiple human-robot interaction studies have demonstrated how robot operators often have inaccurate mental models, resulting in overtrusting the machine, refusing to use it all together, or being unable to keep appropriate focus to monitor and intervene appropriately during a given autonomous task~\cite{strauch:2018}. From the machine side, learning-driven controllers are still opaque with respect to their decision making, causing unpredictable erroneous behaviour in unforeseen environments. 

Mixed initiatives, such as variable autonomy (VA), during a robotic task attempt to solve the above challenges via a dynamic adjustment of control levels between human operators and robots~\cite{methnani:2021,chiou:2021}. 
In a VA system, the robot operator switches from full tele-operation to full autonomy depending on the context and uncertainty associated with the state-action space of the robot. Implementing such a system requires not only sophisticated sensors and algorithms for environmental perception, but also robust communication channels between the human and the robot. Moreover, ethical considerations, such as accountability and transparency, play a crucial role in designing and deploying human-robot systems, ensuring that they are both practical and trusted by their stakeholders~\cite{theodorou_towards_2020}.

Inspired by the dual-process theory \cite{kahneman:2005}, we propose a control architecture for VA systems consisting of a `fast' \textit{(sub)system 1}, which uses deep reinforcement learning~(DRL), and a `slow' \textit{(sub)system 2}, which uses modal logic, working together in a tandem for efficient decision making while taking into consideration its wider environmental and social context.
These two subsystems trigger and provide input to a `negotiation module,' tasked with resolving conflicts and determining the appropriate level of autonomy (LOA). The resulting framework enables the robot and human operator to negotiate the LOA of during operation via an interface of the implemented negotiation model; that is, the negotiation protocol. In particular, we contend that current VA architectures can be further improved by building feedback between the deep learning algorithms that produce the autonomous behavior and the logic-based metacontroller choosing between the human and robot controllers at particular contexts. By using the semantic data that the logic-based metacontroller infers during a particular run to further train the robotic system overcomes the issue of using inaccurate synthetic data, conducting expensive human trials, or tedious data labelling.

\begin{figure}[!t]
    \centering
    \includegraphics[width=\linewidth]{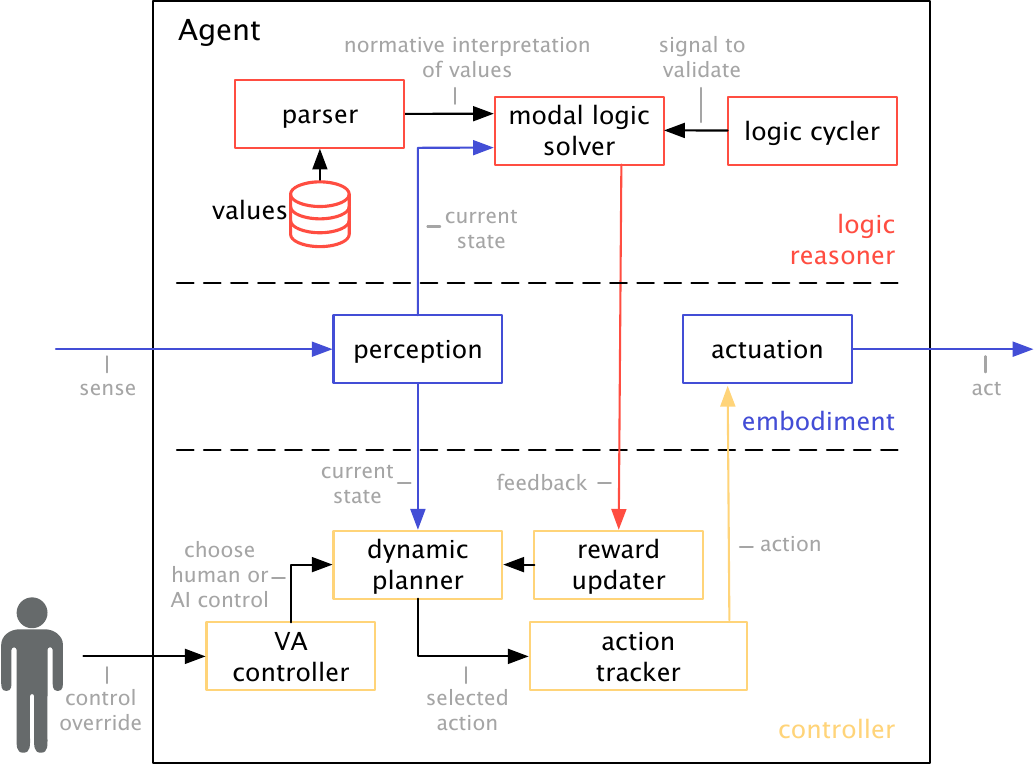}
    \caption{Neurosymbolic VA equips a controller (the `fast' DRL system in yellow) with human value requirements (the `slow' symbolic logic system in red).}
    \label{fig:arch}
\end{figure}

\section{Neurosymbolic variable autonomy}
In this section, we present neurosymbolic VA to developing safe and effective autonomous systems for uncertain environments. By combining the strengths of machine learning and rule-based symbolic reasoning, neurosymbolic VA aims to create well-calibrated autonomous agents that align with human expectations and values. We explore two key aspects of this approach:  dynamic reward structure adjustment in uncertain environments and rule-based symbolic reasoning for safe socio-ethical aligned systems. The proposed architecture (figure~\ref{fig:arch}) uses human data collected during operation to refine reward structures, ensure socio-ethical behavior, and, ultimately, improve the performance and safety of autonomous systems in complex, real-world scenarios.

\subsection{Dynamic reward structure adjustment in uncertain environments}
The `fast' system is a DRL optimiser of desired behaviour. RL algorithms rely on reward structures to optimise desired behaviour. As the reward function explicitly defines the goals and preferences of the system, RL in control \emph{can be} intepretable. By examining the reward function, we can understand which behaviours the agent is pursueing. However, designing a practical reward function can be challenging, as it requires carefully balancing multiple objectives and avoiding unintended consequences in unforeseen and uncertain environments.
In the context of VA, the interpretability of RL allows for incorporating human knowledge and preferences into the learning process. In particular, a plethora of valuable data can be recorded during human control that can improve the system's behaviour in subsequent runs within uncertain environments. Techniques such as cooperative inverse reinforcement learning (CIRL)~\cite{hadfield-menell:2016}, human preference learning~\cite{christiano:2017}, reward shaping~\cite{ng:1999}, safe exploration~\cite{garcia:2012}, and transfer learning~\cite{taylor:2009} can take advantage of this additional data for constructing well-calibrated autonomy.

In a VA system, the operator can provide feedback and improve RL control dynamically through several methods. In CIRL, the human operator's underlying reward function can be estimated by observing their actions and the resulting state transitions. %
Human preference learning allows the RL agent to be rewarded for taking actions similar to those consistently taken by the human operator in specific situations. In simpler cases of reward shaping, higher rewards can be assigned to states and actions that led to successful outcomes in situations where human intervention was necessary. Safe exploration involves identifying regions of the state space considered unsafe or uncertain by analysing the states and actions that prompted the human operator to take control. Transfer learning enables data collected during human operator control to train a separate model that captures the human decision-making process, which can then guide the RL agent's learning process, effectively transferring knowledge from the human operator to the autonomous system. By incorporating these techniques, the reward structure can be refined to align more closely with human judgement, ensuring that the autonomous vehicle behaves consistently with human expectations and values. This well-calibrated autonomy is crucial for building trust and facilitating the system's and human operators' collaboration. %

\subsection{Rule-based symbolic reasoning for interpretable variable autonomy}
The `slow' system is logic-based reasoner giving the means within which human data are interpreted by the DRL controller.
The local logic-based reasoner, acting as a feedback and, if necessary, override mechanism  draws inspiration from Dennett's portrayal of social constructs \cite{dennett_age_2020}.
Using modal logics, the logic-based reasoner validates the socio-ethical behaviour of the agent and provides direction to the reward system of the `fast' system.
The logical formulae interpret the semantics of the systems' actions and then input as further data into the DRL reward structures.
In particular, we provide symbolic interpretations of normative concepts such as socio-ethical values and explicitly desired behaviour.
As the fast system decides to acts, a cycler subcompoment checks if socio-ethical norms are adhered to or not.
Similar to prior work presented in Methnani et al.~\cite{methnani_embracing_2022}, deontic logic can be used to check if the ongoing behaviour of the agent conforms or not the desired normative conduct.

\section{Conclusion}

Neurosymbolic VA is a promising approach to addressing the challenges of developing safe and effective autonomous systems in uncertain environments. By combining the strengths of machine learning and rule-based symbolic reasoning, this framework enables the creation of well-calibrated autonomous agents that align with human expectations and values. The dynamic adjustment of reward structures based on human data collected during operation allows for refining autonomous behaviour, ensuring the system adapts to complex, real-world scenarios. Furthermore, incorporating a logic-based reasoner provides a means to interpret human demonstration data and ensure compliance with socio-ethical norms without requiring a comprehensive understanding of the global system's complexities. Neurosymbolic VA offers a solution to the challenges of balancing autonomy and human intervention,  ultimately paving the way for more reliable, trustworthy, and context-aware autonomous systems.

\bibliography{manuscript,ath}
\bibliographystyle{ieeetr}
\end{document}